\documentclass[sn-mathphys]{sn-jnl}

\jyear{2021}%

\theoremstyle{thmstyleone}%
\theoremstyle{thmstyletwo}%

\theoremstyle{thmstylethree}%

\begin{document}

\title[Deep Leaning-Based Ultra-Fast Stair Detection]{Deep Leaning-Based Ultra-Fast Stair Detection}


\author[1]{\fnm{Chen} \sur{Wang}}\email{venus@buaa.edu.cn}

\author[1]{\fnm{Zhongcai} \sur{Pei}}\email{peizc@buaa.edu.cn}
\equalcont{These authors contributed equally to this work.}

\author[1]{\fnm{Shuang} \sur{Qiu}}\email{zb2003108@buaa.edu.cn}
\equalcont{These authors contributed equally to this work.}

\author*[1]{\fnm{Zhiyong} \sur{Tang}}\email{zyt\_76@buaa.edu.cn}
\equalcont{These authors contributed equally to this work.}

\affil*[1]{\orgdiv{School of Automation Science and Electrical Engineering}, \orgname{Beihang University}, \orgaddress{\city{Beijing}, \postcode{100191}, \country{China}}}


\abstract{Staircases are some of the most common building structures in urban environments. Stair detection is an important task for various applications, including the environmental perception of exoskeleton robots, humanoid robots, and rescue robots and the navigation of visually impaired people. Most existing stair detection algorithms have difficulty dealing with the diversity of stair structure materials, extreme light and serious occlusion. Inspired by human perception, we propose an end-to-end method based on deep learning. Specifically, we treat the process of stair line detection as a multitask involving coarse-grained semantic segmentation and object detection. The input images are divided into cells, and a simple neural network is used to judge whether each cell contains stair lines. For cells containing stair lines, the locations of the stair lines relative to each cell are regressed. Extensive experiments on our dataset show that our method can achieve high performance in terms of both speed and accuracy. A lightweight version can even achieve 300+ frames per second with the same resolution. Our code and dataset will be soon available at GitHub.}

\keywords{Stair detection, Deep leaning, Stair dataset, Group dilated convolution, End-to-end}



\maketitle

\section{Introduction}\label{sec1}

With a long research history in computer vision, stair detection is a fundamental problem and has a wide range of applications. Two kinds of mainstream methods are available for stair detection: line extraction methods \cite{bib1, bib2, bib3, bib4} and plane extraction methods \cite{bib5, bib6, bib7}. For the first type of method, the staircase is defined as a collection of parallel lines. Lines are extracted by applying Canny edge detection, Hough transform and other traditional computer vision algorithms to RGB images or depth images. For the second type of method, the staircase is defined as a collection of parallel planes in three-dimensional space, and the planes are extracted by applying a plane segmentation algorithm to the point cloud data. These two types of methods have long provided relatively reliable stair detection abilities for robots used in urban environments and for visually impaired people. However, there are still some important and challenging problems to be addressed.

The working environment of stair detection determines that a related algorithm usually runs on some small embedded devices. This requires an extremely low computational cost to achieve better real-time performance. To solve this problem, the method used by most algorithms is reducing the input data, namely, reducing the size of the input image or the number of three-dimensional point clouds. For example, in \cite{bib1}, the regions without stairs in the input images are directly discarded by an a priori region of interest (ROI). \cite{bib8} directly removes large planes to reduce the number of input point clouds when scanning the environment.

\begin{figure}[h]%
	\centering
	\includegraphics[width=0.95\textwidth]{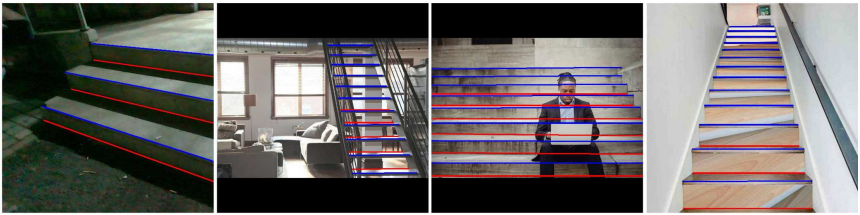}
	\caption{Illustration of the difficulties encountered in stair detection. The convex lines are marked with blue, and the concave lines are marked with red. The figure shows some challenging scenarios, including extreme lighting conditions, special structures, severe occlusions and special materials}\label{fig1}
\end{figure}

The working nature of stair detection determines that a related algorithm should have high reliability and accuracy. The most challenging problem with stair detection is the adaptability of an algorithm to deal with extreme lighting conditions, special structures, severe occlusions and special materials, as shown in Fig. \ref {fig1}. For the method based on line extraction, these problems will be fatal. The reason for this is that the limitations of traditional computer vision based artificial feature extraction approaches make the algorithms difficult to adapt to complex and changeable scenarios. For the method based on plane extraction, the acquisition of point clouds depends on light detection and ranging (LiDAR) or depth cameras, which are not affected by stair texture features and lighting conditions. However, LiDAR and binocular sensors are often expensive and still cannot solve the problem of severe occlusion. 

In addition, when detecting objects with texture features and structures that are similar to stairs, these objects are often misidentified as stairs. The reason for this is that an algorithm based on feature extraction cannot obtain high-level semantic information about stairs like humans, which often leads to false detections and missed detections.

With the above motivations, we propose an end-to-end method based on deep learning that has an extremely fast speed and solves the problem regarding adaptability to different scenarios with monocular vision. Since AlexNet \cite{bib9} established the dominant position of convolutional neural networks (CNNs) in computer vision in 2012, CNNs have rapidly developed in various fields of computer vision due to their strong learning abilities and unique perception modes. The reason for introducing CNNs into stair detection is that an artificial neural network can learn the texture features and high-level semantic information of stairs simultaneously and obtain better robustness by learning datasets that contain various detection scenarios.

Our method is also based on line extraction. The network input is RGB image obtained from visual sensor, and the output is a group of extracted stair lines. All the intermediate processes are computed within the neural network. The key problem of designing this neural network is the representation of staircase features. Approximately two schemes are available for this purpose. 1) The stair detection task is regarded as a semantic segmentation task. The pixels belonging to the stair lines can be taken as positive samples, and the background pixels can be taken as negative samples. However, because the stair lines are usually very thin, the numbers of positive and negative samples will be seriously unbalanced, and the network may have difficulty converging. Additionally, the semantic segmentation framework usually incurs a high computational cost. 2) The stair detection task is regarded as an object detection task, and each stair line is given an external rectangular box. It is easy to know that most boxes will be narrow and long. After experiments, we find that the object detection network has difficulty learning the features of objects.

\begin{figure}[h]%
	\centering
	\includegraphics[width=0.95\textwidth]{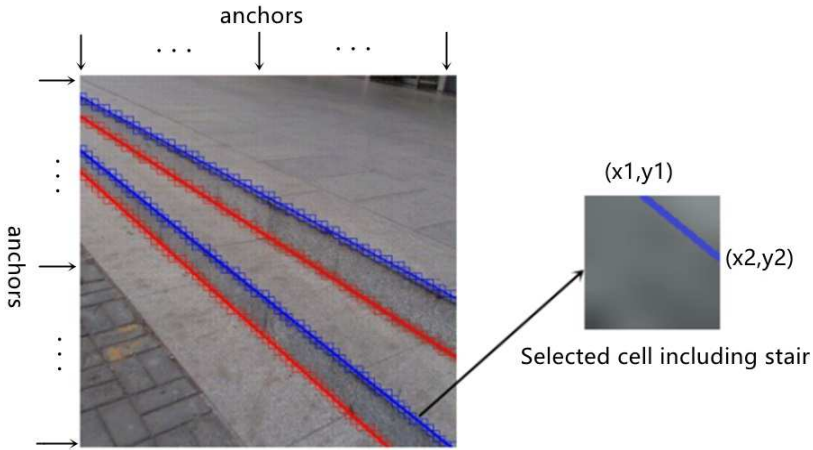}
	\caption{Illustration of a stair feature representation. The whole image on the left is divided into 64x64
		anchors, and the anchors (including stair lines) are detected on the right. The normalized coordinates of the two endpoints relative to the upper left corner of the anchor are regressed}\label{fig2}
\end{figure}

After comprehensively considering these two schemes, we propose a feature representation
method of coarse-grained semantic segmentation combined with object detection, as shown in Fig. \ref {fig2},
which can solve the imbalance between positive and negative samples in scheme 1) and the narrow and
long boxes in scheme 2). As the algorithm is based on coarse-grained segmentation, the size of the final
output feature map is 64x64, which greatly reduces the computational cost relative to that of the
traditional semantic segmentation network. Specifically, we divide the whole input image (input size:
512x512) into 4096 small cells of size 8x8, and a feature map with a size of 64x64 is obtained through
three downsampling operations; then, two heads are connected. One is used for classification to judge
whether the given cell contains convex lines and whether the cell contains concave lines, and the other
is used for location. We regard the cells as the anchors of object detection, and in each anchor, the
normalized coordinates of the stair line relative to the upper left corner of the anchor are regressed. The
two heads work together to detect the whole staircase. In addition, we expand the receptive field by
applying dilated convolution and atrous spatial pyramid pooling (ASPP) \cite{bib10} in the network to improve its perception ability in scenarios without visual clues, such as occlusion and extreme lighting.

In summary, the contribution of this work can be summarized in three parts:

\begin{enumerate}[1.]
\item We provide a stair dataset with fine annotations for stair detection research. The training set contains 2670 images, and the validation set contains 424 images. Each label contains the locations of the two
endpoints and the classification (convex/concave) of each stair line.
	
\item We propose a novel stair detection method based on line extraction. To the best of our knowledge,
this is the first deep learning-based end-to-end stair detection network (called StairNet). Compared with
the line extraction method based on traditional computer vision, our approach not only achieves
extremely fast detection speed but also solves the problem regarding the difficulty of detecting
staircases in challenging scenarios.
	
\item We design a module based on dilated convolution and group convolution. Specifically, we build an
inception \cite{bib11} module by applying group dilated convolution with different dilation rates in the horizontal and
vertical directions, and a channel attention mechanism is also applied so that the network can learn to
extract long-range information features with different aspect ratios.
\end{enumerate}

\section{Related works}\label{sec2}

\subsection{Line extraction methods}\label{subsec2_1}

\subsubsection{Traditional Methods}\label{subsubsec2_1_1}

The main idea of traditional computer vision methods is to use visual clues obtained through edge detection algorithms such as the Canny algorithm, the Sobel filter and line detection algorithms such as the Hough transform \cite{bib12,bib13,bib14}. After performing line extraction, the features of the stairs need to be matched. In \cite{bib15}, the endpoints of the stairs are regarded as three line segments converging at one point for feature extraction. \cite{bib16} regards stair edges arranged in parallel from bottom to top as features, and the upstairs/downstairs labels are classified by a support vector machine (SVM). \cite{bib17} creatively regards the stair structure as a periodic signal in the spatial domain, and its period is the distance between two continuous edges. Then, the 2D fast Fourier transform (FFT) is applied to transform the observed signal to the frequency domain to obtain an image that contains only the edges of stairs. \cite{bib18} proposes a framework based on a unique geometrical feature of a stair. The unique geometrical feature is that every step’s height gradually decreases from the bottom to the top of the staircase. \cite{bib19} proposes a method to identify stairs by using the statistical properties of projection histogram.

\subsubsection{Deep learning methods}\label{subsubsec2_1_2}

Relatively few deep learning computer vision methods are available for stair detection. The main idea of these methods is to extract the ROI containing stairs in the input image through object detection algorithms such as You Only Look Once (YOLO) \cite{bib20} and a region-based CNN (RCNN) \cite{bib21}; then, the traditional computer vision method is applied to extract lines in the ROI \cite{bib22, bib23}. This method divides the stair detection task into two steps, which makes it difficult to ensure real-time performance. In addition, some classification methods utilize deep learning to determine whether an image contains stairs and whether the ROI is upstairs or downstairs \cite{bib24}. Such a classification method does not achieve pixel-level stair localization. It can only be used to provide voice reminders for visually impaired people and not for robot environment perception.

\subsection{Plane extraction methods}\label{subsec2_2}

The main idea of this type of method is to extract potential planes from the input point clouds and filter the planes belonging to stairs in a certain way. Point cloud segmentation is a common plane extraction method, and many methods have been developed for stair feature matching. Classifying planes by obtaining their normal vector and eliminating the planes that do not belong to the stairs is a common method \cite{bib25, bib26}. \cite{bib27} presents an algorithm for stair detection from point clouds based on a new minimal 3D map representation and the estimation of step-like features that are grouped based on adjacency in order to emerge dominant staircase structures. \cite{bib8} proposes a stair plane extraction algorithm based on supervoxel clustering. \cite{bib28} uses the random sample consensus (RANSAC) algorithm to extract planes and then models the corresponding stairs. \cite{bib29} obtains the ground plane through the analysis and processing of point clouds and then detects a group of continuous rising planes as stair features.

\section{Method}\label{sec3}

In this section, we describe the details of our method, including the network architecture of StairNet and the group dilated convolution with different dilation rates in the horizontal and vertical directions. Finally, we briefly introduce the design of the employed loss function.

\subsection{Network architecture}\label{subsec3_1}

\begin{figure}[h]%
	\centering
	\includegraphics[width=0.95\textwidth]{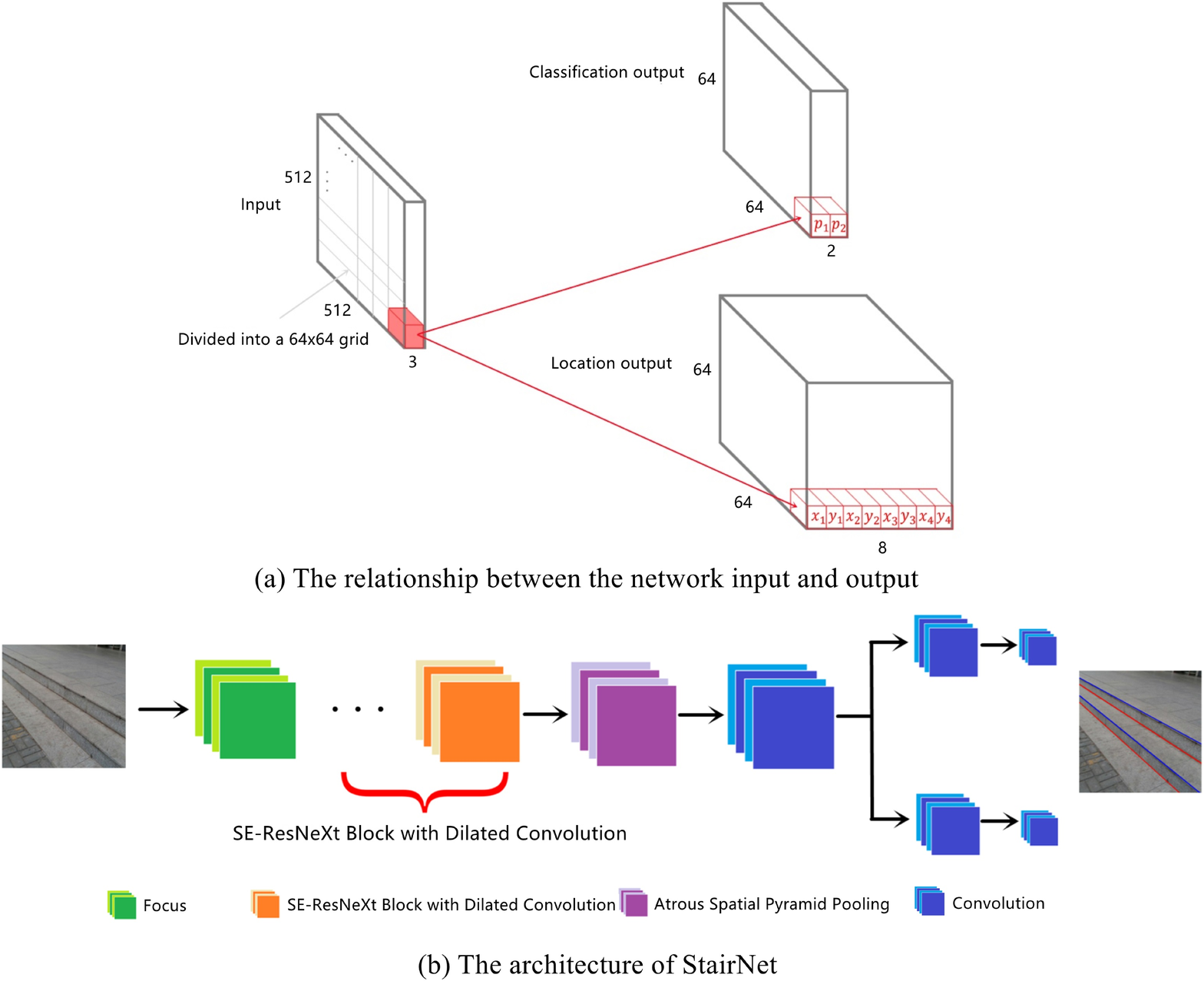}
	\caption{Illustration of the network architecture. Fig. (a) shows the relationship between the network input and output. The network takes a 512x512 full-color image as input, and the output of the network is divided into two branches. The output target values for the classification branch are stored in a 3D tensor of size 64x64x2, which is used to judge whether the cell contains convex lines and whether the cell contains concave lines. The output target values for the location branch are stored in a 3D tensor of size 64x64x8, which is used to predict the locations of two sets of stair lines. Fig. (b) shows the architecture of our network; our backbone contains stacked SE-ResNeXt blocks with dilated convolution, a focus module and an ASPP module}\label{fig3}
\end{figure}

As described in the section \ref{sec1}, we propose a feature representation method involving coarse-grained semantic segmentation combined with object detection for stair detection. Our model takes a 512x512 full-color image as input and processes it with a fully convolutional architecture. A feature map with a size of 64x64 is obtained after three downsampling operations. The output of the network is divided into two branches, and each grid location in the 3D output tensor is associated with a multidimensional vector, as shown in Fig. \ref {fig3}(a).

The output target values for the classification branch are stored in a 3D tensor of size 64x64x2. Considering that a cell may contain both convex lines and concave lines, we study reference \cite{bib30} and use two independent logical classifiers instead of the softmax function to judge whether the given cell contains convex lines and whether the cell contains concave lines.

The output target values for the location branch are stored in a 3D tensor of size 64x64x8, and each cell predicts two sets of locations (x1, y1, x2, y2) and (x3, y3, x4, y4). Regardless of the posture of the stair line in the cell, (x1, y1) and (x3, y3) always represent the location of the left endpoint, and (x2, y2) and (x4, y4) always represent the location of the right endpoint. In addition, the reason for the prediction of two sets of locations is that a few cells will contain two stair lines after image segmentation. For cells with only one stair line, the two sets of locations are given the same label.

As shown in Fig. \ref {fig3}(b), the backbone network consists of a focus module \cite{bib31}, several squeeze-and-excitation (SE)-ResNeXt \cite{bib32, bib33} blocks with dilated convolution and an ASPP module. Each part is introduced in detail below:

\subsubsection{Focus module}\label{sec3_1_1}

In the shallow calculation process of a deep neural network, due to the large size of the input image, downsampling is usually required. Common downsampling methods include pooling, convolution with stride  \textgreater 1 and tensor slicing. Pooling causes the loss of details due to the reduction in resolution, which is unfavorable for segmentation \cite{bib34}. A convolution with stride \textgreater 1 can extract features while downsampling. Tensor slicing retains the original features as much as possible while downsampling. For stair detection, we believe that the texture features of stairs should be kept in the shallow layer of the network. Therefore, we use a focus module as the initial module of the network for downsampling. The focus module is essentially a tensor slicing operation, which is similar to the pass-through layer in \cite{bib35}.

\subsubsection{SE-ResNeXt block}\label{sec3_1_2}

Based on the ResNeXt block presented in \cite{bib33}, we add the channel attention mechanism of SENet and apply dilated convolution to obtain a larger receptive field. ResNeXt uses standard group convolution in its bottleneck, which is also an inception module. ResNeXt with group convolution can reduce the computational cost and achieve higher accuracy than ResNet \cite{bib36}. For the details of our bottleneck, see Section \ref{subsec3_2}. Inspired by ENet \cite{bib37}, standard SE-ResNeXt blocks and SE-ResNeXt blocks with dilated convolution are connected in series to form our backbone. See Table \ref{tab1} for the detailed architecture of the network.

\begin{table}[h]
	\begin{center}
		\begin{minipage}{\textwidth}
			\caption{StairNet architecture}\label{tab1}
			\begin{tabular}{@{}llllll@{}}
				\toprule
				\multicolumn{2}{l}{Name} & \multicolumn{2}{l}{Type} &\multicolumn{2}{l}{Output size}\\
				\midrule
				\multicolumn{2}{l}{Initial} & \multicolumn{2}{l}{Tensor slice} &\multicolumn{2}{l}{256 x 256 x 64}\\
				\midrule
				\multicolumn{2}{l}{Bottleneck 1.0} & \multicolumn{2}{l}{Downsampling} &\multicolumn{2}{l}{128 x 128 x 256}\\
				\multicolumn{2}{l}{Bottleneck 1.1} & \multicolumn{2}{l}{} &\multicolumn{2}{l}{128 x 128 x 256}\\
				\multicolumn{2}{l}{Bottleneck 1.2} & \multicolumn{2}{l}{} &\multicolumn{2}{l}{128 x 128 x 256}\\
				\midrule
				\multicolumn{2}{l}{Bottleneck 2.0} & \multicolumn{2}{l}{Downsampling} &\multicolumn{2}{l}{64 x 64 x 512}\\
				\multicolumn{2}{l}{Bottleneck 2.1} & \multicolumn{2}{l}{Dilated (1,2) and (2,2)} &\multicolumn{2}{l}{64 x 64 x 512}\\
				\multicolumn{2}{l}{Bottleneck 2.2} & \multicolumn{2}{l}{} &\multicolumn{2}{l}{64 x 64 x 512}\\
				\multicolumn{2}{l}{Bottleneck 2.3} & \multicolumn{2}{l}{Dilated (2,4) and (4,4)} &\multicolumn{2}{l}{64 x 64 x 512}\\
				\multicolumn{2}{l}{Bottleneck 2.4} & \multicolumn{2}{l}{} &\multicolumn{2}{l}{64 x 64 x 512}\\
				\multicolumn{2}{l}{Bottleneck 2.5} & \multicolumn{2}{l}{Dilated (3,8) and (8,8)} &\multicolumn{2}{l}{64 x 64 x 512}\\
				\multicolumn{2}{l}{Bottleneck 2.6} & \multicolumn{2}{l}{} &\multicolumn{2}{l}{64 x 64 x 512}\\
				\multicolumn{2}{l}{Bottleneck 2.7} & \multicolumn{2}{l}{Dilated (4,16) and (16,16)} &\multicolumn{2}{l}{64 x 64 x 512}\\
				\midrule
				\multicolumn{6}{l}{Repeat Section 2 Without Downsampling}\\
				\midrule
				\multicolumn{2}{l}{ASPP} & \multicolumn{2}{l}{} &\multicolumn{2}{l}{64 x 64 x 512}\\
				\midrule
				\multicolumn{2}{l}{Conv 3x3} & \multicolumn{2}{l}{} &\multicolumn{2}{l}{64 x 64 x 128}\\
				classification   & location   & classification  & location   & classification   & location  \\
				Conv 3 x 3   & Conv 3 x 3   &   &    & 64 x 64 x 128   & 64 x 64 x 128  \\
				Conv 1 x 1   & Conv 1 x 1   &   &    & 64 x 64 x 2   & 64 x 64 x 8  \\
				& Sigmoid\footnotemark[1]   &   &  Activation  & 64 x 64 x 2   & 64 x 64 x 8  \\
				\botrule
			\end{tabular}
			\footnotetext[1]{The sigmoid activation function is added at the end of the location branch to limit the output to (0,1). This is used to accelerate network training and obtain the normalized coordinates of the stair lines within each cell.}
			
		\end{minipage}
	\end{center}
\end{table}

\subsubsection{ASPP module}\label{sec3_1_3}

\begin{figure}[h]%
	\centering
	\includegraphics[width=0.5\textwidth]{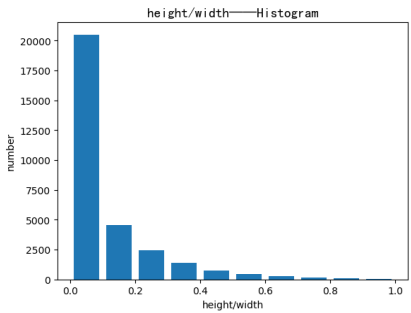}
	\caption{Aspect ratios of all stair lines in the whole dataset. The aspect ratios of most stair lines are within the range of 0--0.2}\label{fig4}
\end{figure}

\begin{figure}[h]%
	\centering
	\includegraphics[width=0.6\textwidth]{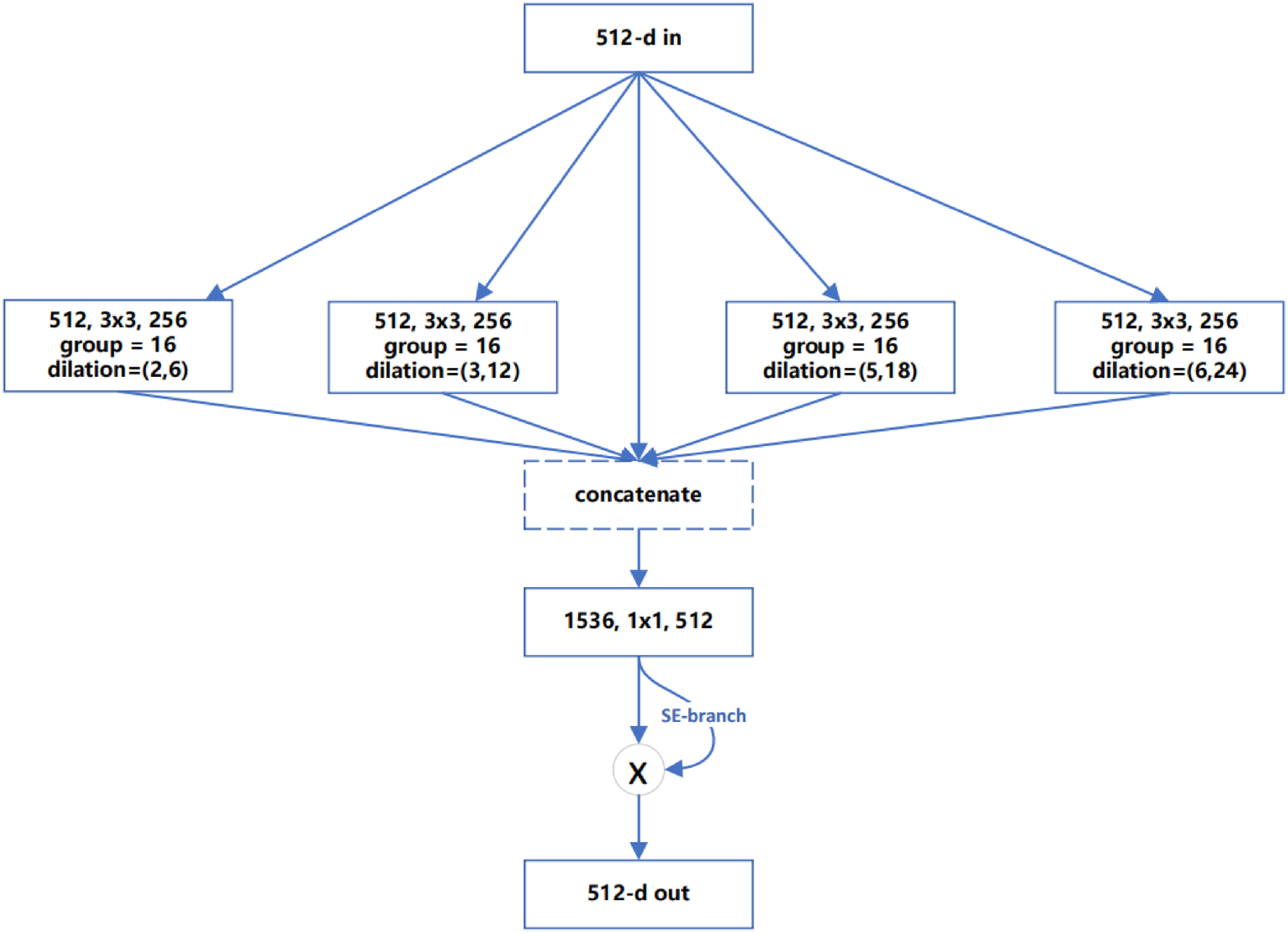}
	\caption{The ASPP module in StairNet: Our ASPP module has five branches, including an input feature map and group dilated convolutions with dilation rates of (2,6), (3,12), (5,18) and (6,24). At the end of the module, a channel attention mechanism is added to learn the emphasis placed on context information extraction at different scales. A layer is denoted as (input channels, filter size, output channels)}\label{fig5}
\end{figure}

To further expand the receptive field and enhance the learning ability of our network with respect to semantic stair information, we apply an ASPP module. In \cite{bib10}, the author applies dilated convolution with dilation rates of (6,6), (12,12), (18,18) and (24,24) to extract features in parallel to capture object and image context information at multiple scales. We apply a module similar to the ASPP module in \cite{bib10} and replace the dilation rates with (2,6), (3,12), (5,18) and (6,24). The reason for applying dilated convolution with different dilation rates in the horizontal and vertical directions is that the distributions of stair lines in the dataset are usually transverse; therefore, dilated convolution with a larger transverse dilation rate is conducive to the detection of stair lines. Specifically, we count the aspect ratios of all stair lines in the whole dataset, and the histogram is shown in Fig. \ref {fig4}. The aspect ratios of most stair lines are within the range of 0--0.2, so we set the aspect ratios of dilation rates at approximately 0.2. In addition, our ASPP module contains a parallel branch for the input feature map and adds a channel attention mechanism to learn the emphasis placed on context information extraction at different scales, as shown in Fig. \ref {fig5}.

\subsection{Dilated group convolution with different dilation rates}\label{subsec3_2}

\begin{figure}[h]%
	\centering
	\includegraphics[width=0.6\textwidth]{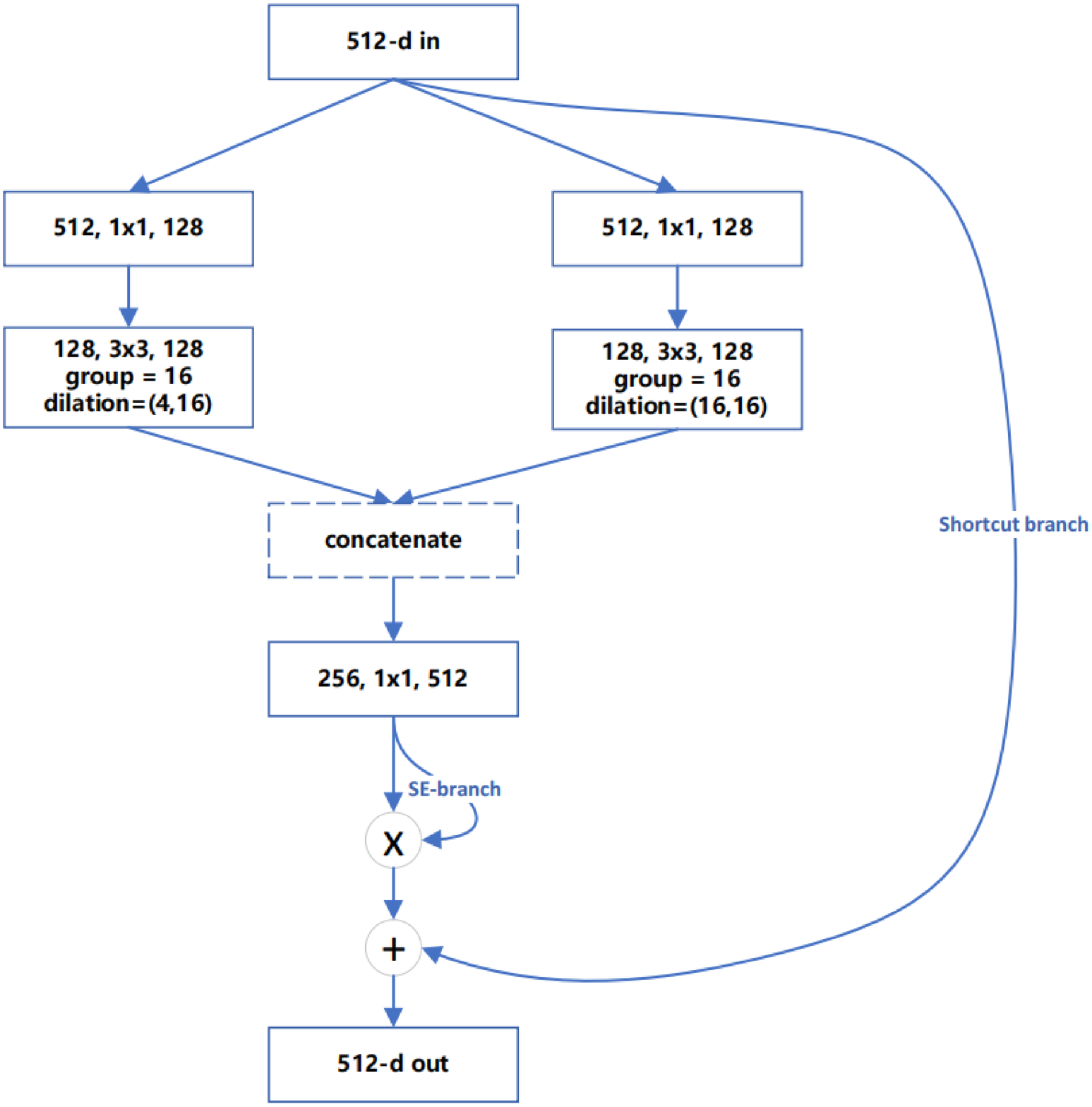}
	\caption{Illustration of the SE-ResNeXt block with dilated convolution. The original standard group convolution with 32 groups is divided into two branches with 16 groups, and the dilation rates of the dilated convolutions are (4, 16) and (16, 16). Finally, the results of the two branches are concatenated. A layer is denoted as (input channels, filter size, output channels)}\label{fig6}
\end{figure}

As mentioned in Section \ref{sec3_1_2}, to increase the receptive field and improve the segmentation performance of our network, we apply dilated convolution in the SE-ResNeXt block. While applying dilated convolution, we redesign the standard group convolution of the ResNeXt block. We build an inception module by applying group dilated convolutions with different dilation rates in the horizontal and vertical directions. Specifically, we concatenate the calculation results of standard dilated convolutions and the calculation results of dilated convolutions with different dilation rates in the horizontal and vertical directions, and then we apply a channel attention mechanism to learn the weights of the concatenated channels. Fig. \ref {fig6} shows the structure of Bottleneck 2.7 in Table \ref{tab1}, and other bottlenecks can be obtained in the same way.

Similar to the dilation rates in the ASPP module in Section \ref{sec3_1_3}, we consider that the distributions of stairs in the dataset are usually transverse. To use this prior knowledge, we calculate asymmetric dilated convolutions with dilation rates of (1,2), (2,4), (3,8), and (4,16) and their corresponding standard dilated convolutions and concatenate the results. An asymmetric dilated convolution is helpful for learning the features of a single stair line, while the standard dilated convolution is helpful for learning the contextual features between stair lines.

\subsection{Loss function}\label{sec3_3}

Stair detection is a typical multitask of classification and regression. Our loss function inherits the multitask loss idea used for most object detection tasks. The loss function includes a classification loss and a location loss. The specific formula is as follows:

\begin{equation}
	L(\{p_{ij}\},\{t_{ij}\})=\frac{1}{N^{2}}(\sum_{i}^{N}\sum_{j}^{N}L_{cls}(p_{ij},p_{ij}^*)+\lambda\sum_{i}^{N}\sum_{j}^{N}p_{ij}L_{loc}(t_{ij},t_{ij}^*))\qquad\label{eq1}
\end{equation}
where N represents the number of cells in a row or column, namely, 64, and i and j represent the position of the cell in the whole image. $ p_{ij} $ is a 2-dimensional vector that indicates the prediction probability regarding whether the cell contains convex lines and whether the cell contains concave lines. The corresponding ground truth $ p_{ij}^* $ has four values: (1,0), (1,1), (0,1) and (0,0), which represent only convex lines, both convex lines and concave lines, only concave lines and no lines, respectively. $ t_{ij} $ is an 8-dimensional vector, which represents the normalized coordinates of the two sets of locations predicted by the cell; $ t_{ij}^* $ is the corresponding ground truth; $ \lambda $ is the weight coefficient, which is set to 4 here. Since we only calculate cells that contain stair lines, we use the vector $ p_{ij} $ dot vector $ L_{loc} $ with the broadcasting mechanism of PyTorch.

For the classification loss $ L_{cls} $, the binary cross-entropy loss function with sigmoid activation is applied to judge whether the given cell contains lines. For the location loss $ L_{loc} $, the mean square loss function is applied. According to the prior knowledge that stairs are usually distributed horizontally in an image, we need to strengthen the localization of the ordinate. Therefore, the location loss is divided into two parts according to the abscissa and ordinate, and these parts are given different weights; that is: 

\begin{equation}
	L_{loc}(t_{ij},t_{ij}^*)=L_{loc}(x_{ij},x_{ij}^*)+\alpha L_{loc}(y_{ij},y_{ij}^*)\qquad\qquad\qquad\qquad\qquad\qquad\label{eq2}
\end{equation}
where $ x_{ij} $ is a 4-dimensional vector that represents the 4 abscissa values predicted by the cell and $ x_{ij}^* $ is the corresponding ground truth. $ y_{ij} $ is a 4-dimensional vector that represents the 4 ordinate values predicted by the cell, and $ y_{ij}^* $ is the corresponding ground truth. $ \alpha $ is the weight coefficient, which is set to 4 here.

\section{Experiment}\label{sec4}

In this section, we describe the details of the conducted experiment, including the introduction of the dataset, the model evaluation method, the training strategy, an ablation experiment and performance testing on the dataset.

\subsection{Experimental settings}\label{sec4_1}

\subsubsection{Dataset introduction}\label{sec4_1_1}

\begin{figure}[h]%
	\centering
	\includegraphics[width=0.95\textwidth]{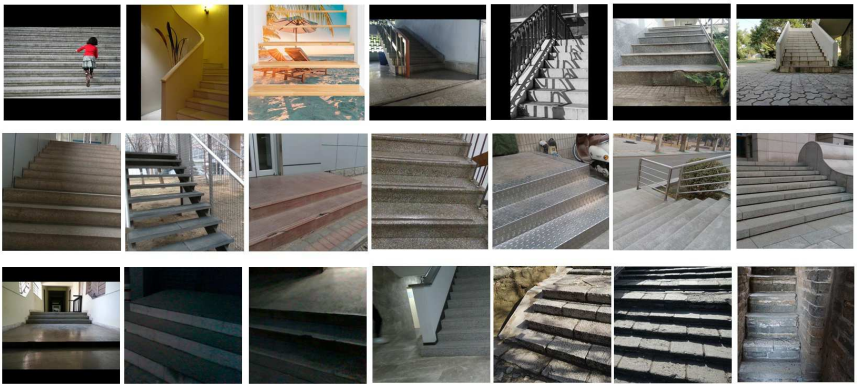}
	\caption{Partial images found in the training set}\label{fig7}
\end{figure}

The main sources of our dataset are as follows. First, we relabel the dataset of reference \cite{bib22} and add it to our dataset. Then, we use a camera to collect stair images from actual scenes at Beihang University, as well as a few stair images from the Great Wall. Finally, we download some stair images from the internet. These images are padded and resized to 512x512 to simplify the data loading process. The whole dataset contains a total of 3094 images, which are randomly divided into 2670 images for the training set and 424 images for the validation set. Fig. \ref {fig7} shows some images in the training set.

The annotation form of the dataset is as follows: 

cls x1 y1 x2 y2/n 

... 

Each stair line is represented by the above five-tuple data, where cls represents the class of the stair line, 0 represents a convex line and 1 represents a concave line. X1 and y1 represent the coordinates of the left endpoint of the stair line, and x2 and y2 represent the coordinates of the right endpoint of the stair line. The label of an image is stored in a text file and associated by the file name.

\subsubsection{Training strategy}\label{sec4_1_2}

We train the model on a workstation with an i7-9700 GPU and an RTX 3080 GPU by using the PyTorch framework. As mentioned above, the input size of the network is 512x512, training is conducted for a total of 200 epochs, and the batch size is set to 4. The Adam optimizer is used, the weight decay is set to $ 10^{-6} $, and the initial learning rate is set to 0.0005. In addition, we apply a dynamic learning rate adjustment strategy, where the learning rate is halved every 50 epochs.

In terms of data enhancement, we mainly use a random mirror with a probability of 0.5 and random occlusion with a probability of 0.5. Random mirror is used to eliminate the uneven distribution of ROIs in the training images. Random occlusion is used to simulate the situation in which the stairs are often blocked by pedestrians and other objects in reality

\subsubsection{Evaluation metrics}\label{sec4_1_3}

In essence, our task is still an object detection problem based on coarse-grained segmentation. There are only four kinds of cells in our task, namely, cells with only convex lines, cells with only concave lines, cells with both kinds of lines and cells with no lines. Since the background cells are easy to classify and a few cells with both lines are difficult to locate, to objectively evaluate the performance of the model, we use the frequency weighted intersection over union (FWIOU) \cite{bib38} as the evaluation method, and the background class is not calculated.

\begin{equation}
	FWIOU=\frac{1}{\sum\limits_{i=0}^{k}\sum\limits_{j=0}^{k}p_{ij}}\sum_{i=0}^{k}\frac{\sum\limits_{j=0}^{k}p_{ij}p_{ii}}{\sum\limits_{j=0}^{k}p_{ij}+\sum\limits_{j=0}^{k}p_{ji}-p_{ii}}\qquad\qquad\qquad\qquad\qquad\label{eq3}
\end{equation}
where $ p_{ij} $ is the number of pixels of class i inferred to belong to class j. Namely, $ p_{ii} $ represents the number of true positives (TP), while $ p_{ij} $ and $ p_{ji} $ are usually interpreted as false positives (FP) and false negatives (FN), respectively. Then, the above formula can be rewritten as follows:

\begin{equation}
 	FWIOU=\frac{1}{\sum\limits_{i=0}^{k}\sum\limits_{j=0}^{k}p_{ij}}\sum_{i=0}^{k}\frac{\sum\limits_{j=0}^{k}p_{ij}TP}{TP+FP+FN}\qquad\qquad\qquad\qquad\qquad\qquad\label{eq4}
\end{equation}

When applying equation (\ref{eq4}), $ p_{ij} $ is regarded as the number of cells of class i inferred to belong to class j. The judgment of true positives and false positives depends not only on the classes of cells but also on the locations of lines in these cells. In other words, a TP cell must meet the following two conditions: 1) the cell is a positive sample and is correctly predicted as a positive sample; 2) in the cell, the location error between the predicted location of the line and the corresponding ground truth is within a certain threshold.

In the object detection task, the intersection over union(IOU) is used to measure the proximity between two boxes. Similarly, we need an index to measure the proximity of two line segments. The location of a line segment is determined by its two endpoints, so the problem can be transformed into measuring the proximity between endpoints. Inspired by reference \cite{bib39}, we apply equation (\ref{eq5}) to convert the distance between two endpoints into a confidence score.

\begin{equation}
	c(x)=
	\begin{cases}
	
	\frac{e^{\alpha(1-\frac{D_T(x)}{d_{th}})}-1}{e^{\alpha}-1}, & if\,D_T(x)\le d_{th}\qquad\qquad\qquad\qquad\qquad\qquad\quad\ \\
	0,& \text{otherwise}
	
	\end{cases}\label{eq5}
\end{equation}
where c(x) is the confidence and $ D_T(x) $ is defined as the 2D Euclidean distance in the image space. $ d_{th} $ is the distance threshold and is set to 1. The sharpness of the exponential function is defined by the parameter $ \alpha $. To achieve precise localization with this function, $ \alpha $ is set to 2. In practice, since a line segment has no direction, we apply equation (\ref{eq5}) to an endpoint on one line segment and the endpoint closest to it on the other line segment. After obtaining the confidence of the two endpoints, we calculate the mean value and assign it as the final confidence score.

In the following content, we use the accuracy, recall and FWIOU metrics when c(x)=0.5, as well as the mean FWIOU (mFWIOU), as the evaluation indicators of the model. The mFWIOU is defined by the mean value of the FWIOUs obtained under 19 confidence values when c(x)=0.05--0.95 with a step size of 0.05.

\subsection{Ablation experiments}\label{sec4_2}

In this section, we verify our method with several ablation studies. The experiments are all conducted with the same settings as those described in Section \ref{sec4_1}. We take the network stacked with standard SE-ResNeXt blocks as the baseline. Based on this, we study the influence of the focus module, ASPP module and SE-ResNeXt blocks with dilated convolution on the performance of the model. The experimental results are shown in the table below.

\begin{table}[h]
	\begin{center}
		\begin{minipage}{\textwidth}
			\caption{Results of ablation experiments}\label{tab2}
			\setlength{\tabcolsep}{1mm}{
			\begin{tabular}{@{}llllllll@{}}
				\toprule
				Backbone  & Dilation & Focus  & ASPP  & Accuracy   & Recall & FWIOU & mFWIOU\\
				\midrule
				SE-ResNeXt (baseline)   &  &   &   & 79.96\% & 81.41\%   & 67.83\%  & 58.13\% \\
				SE-ResNeXt + dilation   & \checkmark   &   &  & 81.07\%  & 81.92\%   & 69.00\%  & 59.11\% \\
				SE-ResNeXt + focus      &   &\checkmark   &   & 80.56\%  & 80.90\%   & 67.91\%  & 58.07\% \\
				SE-ResNeXt + ASPP       &   &   &\checkmark   & 81.27\%  & 80.93\%   & 68.45\%  & 58.57\% \\
				SE-ResNeXt + dilation + focus   &\checkmark   &\checkmark   &   & 81.47\%  & 81.68\%   & 69.11\%  & 59.32\% \\
				SE-ResNeXt + focus + ASPP       &   &\checkmark   &\checkmark   & 81.24\%  & 81.32\%   & 68.69\%  & 58.79\% \\
				SE-ResNeXt + dilation + ASPP    &\checkmark   &   &\checkmark   & 81.47\%  & 81.13\%   & 68.71\%  & 59.06\% \\
				StairNet    &\checkmark   &\checkmark   &\checkmark   & 81.49\%  & 81.91\%   & 69.31\%  & 59.51\% \\
				\botrule
			\end{tabular}}
		\end{minipage}
	\end{center}
\end{table}

The results show that the SE-ResNeXt blocks with dilated convolution can significantly improve the performance of the model. In addition, the focus module and ASPP module slightly improve the performance of the model.

\subsection{Performance experiments}\label{sec4_3}

In this section, we present model performance experiments conducted on the validation set, which are mainly performed to determine the inference speed and accuracy of the model. We provide three versions of StairNet, including StairNet 1x, StairNet 0.5x and StairNet 0.25x, to meet the requirements of devices with different computation capabilities. The size of the model is scaled by a channel width factor. We test the three versions on a desktop platform and an embedded platform. The specific experimental platforms, model sizes and inference speeds are shown in Table \ref {tab3}.

\begin{table}[h]
	\begin{center}
		\begin{minipage}{\textwidth}
			\caption{Results of the model inference speed experiment}\label{tab3}
			\setlength{\tabcolsep}{1mm}{
				\begin{tabular}{@{}llll@{}}
					\toprule
					Platform  & StairNet 1x(35.1Mb) & StairNet 0.5x(9.1Mb)  & StairNet 0.25x(2.48Mb)\\
					\midrule
					i7-9700 + RTX 3080   & 12.48ms & 5.92ms   & 3.07ms  \\
					NVIDIA Jetson NX   & 219.07ms   & 97.45ms  & 42.31ms\\
					\botrule
			\end{tabular}}
		\end{minipage}
	\end{center}
\end{table}

The results show that our three models can meet the real-time requirements of the desktop platform, and the 0.5x and 0.25x models can meet the real-time requirements of the embedded platform.

To objectively evaluate the accuracy of the model, we divide the data in the validation set into daytime data, night data and network data according to their collection conditions, and the detection difficulty also increases in sequence. The accuracy results are shown in Table \ref {tab4}.

\begin{table}[h]
	\begin{center}
		\begin{minipage}{\textwidth}
			\caption{Results of the model accuracy experiment}\label{tab4}
			\setlength{\tabcolsep}{0.7mm}{
			\begin{tabular*}{\textwidth}{@{}lllllllllllll@{}}
				\toprule
				& \multicolumn{3}{@{}l@{}}{Accuracy(\%)} & \multicolumn{3}{@{}l@{}}{Recall(\%)} & \multicolumn{3}{@{}l@{}}{FWIOU(\%)} & \multicolumn{3}{@{}l@{}}{mFWIOU(\%)} \\\cmidrule{2-4}\cmidrule{5-7}\cmidrule{8-10}\cmidrule{11-13}
				Model & Day & Night & Net & Day & Night & Net & Day & Night & Net & Day & Night & Net \\
				\midrule
			    StairNet 1x  & 86.29 &79.24 &72.21 &87.05 &83.43 &68.94 &76.65 &68.61 &54.97 & 66.00 & 58.91 & 46.78\\
			    StairNet 0.5x &85.32 &78.81 &71.00 &86.55 &83.29 &68.53 &75.53 &68.17 &53.97 &64.78 &57.79 &45.71\\
			    StairNet 0.25x &82.84 &75.52 &67.80 &84.73 &81.20 &66.96 &72.28 &64.38 &51.21 &62.03 &55.17 &43.40\\
				\botrule
			\end{tabular*}}
		\end{minipage}
	\end{center}
\end{table}

The results show that the performance of the 0.5x model is slightly worse than that of the 1x model, but the performance of the 0.25x model is much lower than that of the 0.5x and 1x models. For all models, the daytime data detection accuracy is greater than the night and network data detection accuracies.

\begin{figure}[h]%
	\centering
	\includegraphics[width=0.95\textwidth]{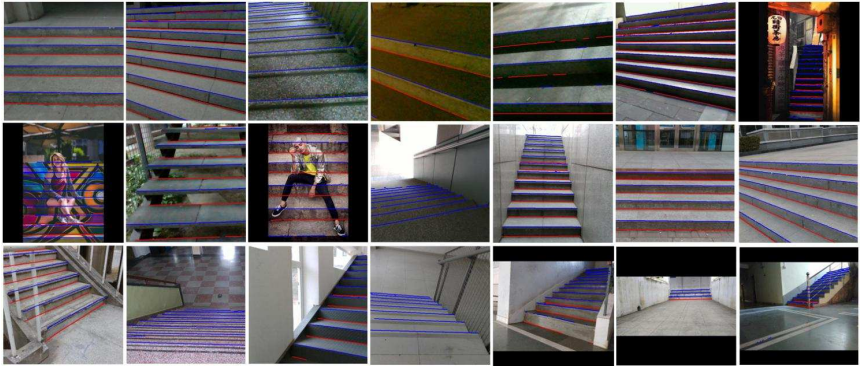}
	\caption{Partial visualization results produced by StairNet 1x on the validation set}\label{fig8}
\end{figure}

Fig. \ref {fig8} shows some visualization results produced by StairNet 1x on the validation set. These staircases have different building structures, materials, shooting angles and lighting conditions. Our method can still obtain satisfactory results under the conditions of extreme lighting, serious occlusion and special stair structures and materials.

\section{Conclusion}\label{sec5}

We propose a novel fully convolutional neural network architecture that regards the stair detection task as a combination of semantic segmentation and object detection, where the aim is to quickly and accurately detect stair lines in monocular vision in an end-to-end manner. In addition, we provide a dataset with fine annotations for stair detection research. Experiments conducted on this dataset
demonstrate the effectiveness of our method. Finally, experiments performed on a Jetson NX show that our model can run in real time on an embedded device. It can effectively make use of the computing resources provided by the embedded platform and serve as an edge computing solution for stair detection in various devices.

\bibliography{Deep_Leaning-Based_Ultra-Fast_Stair_Detection}


\begin{thebibliography}{39}
\ifx \bisbn   \undefined \def \bisbn  #1{ISBN #1}\fi
\ifx \binits  \undefined \def \binits#1{#1}\fi
\ifx \bauthor  \undefined \def \bauthor#1{#1}\fi
\ifx \batitle  \undefined \def \batitle#1{#1}\fi
\ifx \bjtitle  \undefined \def \bjtitle#1{#1}\fi
\ifx \bvolume  \undefined \def \bvolume#1{\textbf{#1}}\fi
\ifx \byear  \undefined \def \byear#1{#1}\fi
\ifx \bissue  \undefined \def \bissue#1{#1}\fi
\ifx \bfpage  \undefined \def \bfpage#1{#1}\fi
\ifx \blpage  \undefined \def \blpage #1{#1}\fi
\ifx \burl  \undefined \def \burl#1{\textsf{#1}}\fi
\ifx \doiurl  \undefined \def \doiurl#1{\url{https://doi.org/#1}}\fi
\ifx \betal  \undefined \def \betal{\textit{et al.}}\fi
\ifx \binstitute  \undefined \def \binstitute#1{#1}\fi
\ifx \binstitutionaled  \undefined \def \binstitutionaled#1{#1}\fi
\ifx \bctitle  \undefined \def \bctitle#1{#1}\fi
\ifx \beditor  \undefined \def \beditor#1{#1}\fi
\ifx \bpublisher  \undefined \def \bpublisher#1{#1}\fi
\ifx \bbtitle  \undefined \def \bbtitle#1{#1}\fi
\ifx \bedition  \undefined \def \bedition#1{#1}\fi
\ifx \bseriesno  \undefined \def \bseriesno#1{#1}\fi
\ifx \blocation  \undefined \def \blocation#1{#1}\fi
\ifx \bsertitle  \undefined \def \bsertitle#1{#1}\fi
\ifx \bsnm \undefined \def \bsnm#1{#1}\fi
\ifx \bsuffix \undefined \def \bsuffix#1{#1}\fi
\ifx \bparticle \undefined \def \bparticle#1{#1}\fi
\ifx \barticle \undefined \def \barticle#1{#1}\fi
\bibcommenthead
\ifx \bconfdate \undefined \def \bconfdate #1{#1}\fi
\ifx \botherref \undefined \def \botherref #1{#1}\fi
\ifx \url \undefined \def \url#1{\textsf{#1}}\fi
\ifx \bchapter \undefined \def \bchapter#1{#1}\fi
\ifx \bbook \undefined \def \bbook#1{#1}\fi
\ifx \bcomment \undefined \def \bcomment#1{#1}\fi
\ifx \oauthor \undefined \def \oauthor#1{#1}\fi
\ifx \citeauthoryear \undefined \def \citeauthoryear#1{#1}\fi
\ifx \endbibitem  \undefined \def \endbibitem {}\fi
\ifx \bconflocation  \undefined \def \bconflocation#1{#1}\fi
\ifx \arxivurl  \undefined \def \arxivurl#1{\textsf{#1}}\fi
\csname PreBibitemsHook\endcsname

\bibitem{bib1}
\begin{bchapter}
\bauthor{\bsnm{Shahrabadi}, \binits{S.}},
\bauthor{\bsnm{Rodrigues}, \binits{J.M.}},
\bauthor{\bsnm{Buf}, \binits{J.}}:
\bctitle{Detection of indoor and outdoor stairs}.
In: \bbtitle{Iberian Conference on Pattern Recognition \& Image Analysis},
pp. \bfpage{847}--\blpage{854}
(\byear{2013}).
\doiurl{10.1007/978-3-642-38628-2_100}
\end{bchapter}
\endbibitem

\bibitem{bib2}
\begin{barticle}
\bauthor{\bsnm{Wang}, \binits{S.}},
\bauthor{\bsnm{Pan}, \binits{H.}},
\bauthor{\bsnm{Zhang}, \binits{C.}},
\bauthor{\bsnm{Tian}, \binits{Y.}}:
\batitle{Rgb-d image-based detection of stairs, pedestrian crosswalks and
  traffic signs}.
\bjtitle{Journal of Visual Communication \& Image Representation}
\bvolume{25}(\bissue{2}),
\bfpage{263}--\blpage{272}
(\byear{2014}).
\doiurl{10.1016/j.jvcir.2013.11.005}
\end{barticle}
\endbibitem

\bibitem{bib3}
\begin{bchapter}
\bauthor{\bsnm{Krausz}, \binits{N.E.}},
\bauthor{\bsnm{Hargrove}, \binits{L.J.}}:
\bctitle{Recognition of ascending stairs from 2d images for control of powered
  lower limb prostheses}.
In: \bbtitle{2015 7th International IEEE/EMBS Conference on Neural Engineering
  (NER)},
pp. \bfpage{615}--\blpage{618}
(\byear{2015}).
\doiurl{10.1109/NER.2015.7146698}
\end{bchapter}
\endbibitem

\bibitem{bib4}
\begin{bchapter}
\bauthor{\bsnm{Wang}, \binits{S.}},
\bauthor{\bsnm{Tian}, \binits{Y.}}:
\bctitle{Detecting stairs and pedestrian crosswalks for the blind by rgbd
  camera}.
In: \bbtitle{2012 IEEE International Conference on Bioinformatics and
  Biomedicine Workshops},
pp. \bfpage{732}--\blpage{739}
(\byear{2012}).
\doiurl{10.1109/BIBMW.2012.6470227}
\end{bchapter}
\endbibitem

\bibitem{bib5}
\begin{bchapter}
\bauthor{\bsnm{Westfechtel}, \binits{T.}},
\bauthor{\bsnm{Ohno}, \binits{K.}},
\bauthor{\bsnm{Mertsching}, \binits{B.}},
\bauthor{\bsnm{Nickchen}, \binits{D.}},
\bauthor{\bsnm{Kojima}, \binits{S.}},
\bauthor{\bsnm{Tadokoro}, \binits{S.}}:
\bctitle{3d graph based stairway detection and localization for mobile robots}.
In: \bbtitle{2016 IEEE/RSJ International Conference on Intelligent Robots and
  Systems (IROS)},
pp. \bfpage{473}--\blpage{479}
(\byear{2016}).
\doiurl{10.1109/IROS.2016.7759096}
\end{bchapter}
\endbibitem

\bibitem{bib6}
\begin{bchapter}
\bauthor{\bsnm{Zhao}, \binits{X.}},
\bauthor{\bsnm{Chen}, \binits{W.}},
\bauthor{\bsnm{Yan}, \binits{X.}},
\bauthor{\bsnm{Wang}, \binits{J.}},
\bauthor{\bsnm{Wu}, \binits{X.}}:
\bctitle{Real-time stairs geometric parameters estimation for lower limb
  rehabilitation exoskeleton}.
In: \bbtitle{2018 Chinese Control And Decision Conference (CCDC)},
pp. \bfpage{5018}--\blpage{5023}
(\byear{2018}).
\doiurl{10.1109/CCDC.2018.8408001}
\end{bchapter}
\endbibitem

\bibitem{bib7}
\begin{bchapter}
\bauthor{\bsnm{Zheng}, \binits{Z.}},
\bauthor{\bsnm{Zhong}, \binits{G.}},
\bauthor{\bsnm{Deng}, \binits{H.}}:
\bctitle{A method to detect stairs with three-dimensional scanning for hexapod
  robot stair climbing}.
In: \bbtitle{2016 IEEE International Conference on Mechatronics and
  Automation},
pp. \bfpage{2541}--\blpage{2546}
(\byear{2016}).
\doiurl{10.1109/ICMA.2016.7558966}
\end{bchapter}
\endbibitem

\bibitem{bib8}
\begin{barticle}
\bauthor{\bsnm{Oh}, \binits{K.W.}},
\bauthor{\bsnm{Choi}, \binits{K.S.}}:
\batitle{Supervoxel-based staircase detection from range data}.
\bjtitle{IEIE Transactions on Smart Processing \& Computing}
\bvolume{4}(\bissue{6}),
\bfpage{403}--\blpage{406}
(\byear{2015}).
\doiurl{10.5573/IEIESPC.2015.4.6.403}
\end{barticle}
\endbibitem

\bibitem{bib9}
\begin{barticle}
\bauthor{\bsnm{Krizhevsky}, \binits{A.}},
\bauthor{\bsnm{Sutskever}, \binits{I.}},
\bauthor{\bsnm{Hinton}, \binits{G.}}:
\batitle{Imagenet classification with deep convolutional neural networks}.
\bjtitle{Communication of the ACM}
\bvolume{60}(\bissue{6}),
\bfpage{84}--\blpage{90}
(\byear{2017}).
\doiurl{10.1145/3065386}
\end{barticle}
\endbibitem

\bibitem{bib10}
\begin{barticle}
\bauthor{\bsnm{Chen}, \binits{L.-C.}},
\bauthor{\bsnm{Papandreou}, \binits{G.}},
\bauthor{\bsnm{Kokkinos}, \binits{I.}},
\bauthor{\bsnm{Murphy}, \binits{K.}},
\bauthor{\bsnm{Yuille}, \binits{A.L.}}:
\batitle{Deeplab: Semantic image segmentation with deep convolutional nets,
  atrous convolution, and fully connected crfs}.
\bjtitle{IEEE Transactions on Pattern Analysis and Machine Intelligence}
\bvolume{40}(\bissue{4}),
\bfpage{834}--\blpage{848}
(\byear{2018}).
\doiurl{10.1109/TPAMI.2017.2699184}
\end{barticle}
\endbibitem

\bibitem{bib11}
\begin{bchapter}
\bauthor{\bsnm{Szegedy}, \binits{C.}},
\bauthor{\bsnm{Liu}, \binits{W.}},
\bauthor{\bsnm{Jia}, \binits{Y.}},
\bauthor{\bsnm{Sermanet}, \binits{P.}},
\bauthor{\bsnm{Reed}, \binits{S.}},
\bauthor{\bsnm{Anguelov}, \binits{D.}},
\bauthor{\bsnm{Erhan}, \binits{D.}},
\bauthor{\bsnm{Vanhoucke}, \binits{V.}},
\bauthor{\bsnm{Rabinovich}, \binits{A.}}:
\bctitle{Going deeper with convolutions}.
In: \bbtitle{2015 IEEE Conference on Computer Vision and Pattern Recognition
  (CVPR)},
pp. \bfpage{1}--\blpage{9}
(\byear{2015}).
\doiurl{10.1109/CVPR.2015.7298594}
\end{bchapter}
\endbibitem

\bibitem{bib12}
\begin{barticle}
\bauthor{\bsnm{Vu}, \binits{H.}},
\bauthor{\bsnm{Hoang}, \binits{V.-N.}},
\bauthor{\bsnm{Le}, \binits{T.-L.}},
\bauthor{\bsnm{Tran}, \binits{T.-H.}},
\bauthor{\bsnm{Nguyen}, \binits{T.T.}}:
\batitle{A projective chirp based stair representation and detection from
  monocular images and its application for the visually impaired}.
\bjtitle{Pattern Recognition Letters}
\bvolume{137},
\bfpage{17}--\blpage{26}
(\byear{2020}).
\doiurl{10.1016/j.patrec.2019.03.007}
\end{barticle}
\endbibitem

\bibitem{bib13}
\begin{bchapter}
\bauthor{\bsnm{Murakami}, \binits{S.}},
\bauthor{\bsnm{Shimakawa}, \binits{M.}},
\bauthor{\bsnm{Kivota}, \binits{K.}},
\bauthor{\bsnm{Kato}, \binits{T.}}:
\bctitle{Study on stairs detection using rgb-depth images}.
In: \bbtitle{2014 Joint 7th International Conference on Soft Computing and
  Intelligent Systems (SCIS) and 15th International Symposium on Advanced
  Intelligent Systems (ISIS)},
pp. \bfpage{1186}--\blpage{1191}
(\byear{2014}).
\doiurl{10.1109/SCIS-ISIS.2014.7044705}
\end{bchapter}
\endbibitem

\bibitem{bib14}
\begin{bchapter}
\bauthor{\bsnm{Yu}, \binits{S.-H.}},
\bauthor{\bsnm{Yang}, \binits{B.-R.}},
\bauthor{\bsnm{Lee}, \binits{H.-H.}},
\bauthor{\bsnm{Tanaka}, \binits{E.}}:
\bctitle{A ground-stair walking strategy of the assistive device based on the
  rgb-d camera}.
In: \bbtitle{2021 IEEE/SICE International Symposium on System Integration
  (SII)},
pp. \bfpage{341}--\blpage{346}
(\byear{2021}).
\doiurl{10.1109/IEEECONF49454.2021.9382668}
\end{bchapter}
\endbibitem

\bibitem{bib15}
\begin{bchapter}
\bauthor{\bsnm{Khaliluzzaman}, \binits{M.}},
\bauthor{\bsnm{Deb}, \binits{K.}},
\bauthor{\bsnm{Jo}, \binits{K.-H.}}:
\bctitle{Stairways detection and distance estimation approach based on three
  connected point and triangular similarity}.
In: \bbtitle{2016 9th International Conference on Human System Interactions
  (HSI)},
pp. \bfpage{330}--\blpage{336}
(\byear{2016}).
\doiurl{10.1109/HSI.2016.7529653}
\end{bchapter}
\endbibitem

\bibitem{bib16}
\begin{bchapter}
\bauthor{\bsnm{Khaliluzzaman}, \binits{M.}},
\bauthor{\bsnm{Yakub}, \binits{M.}},
\bauthor{\bsnm{Chakraborty}, \binits{N.}}:
\bctitle{Comparative analysis of stairways detection based on rgb and rgb-d
  image}.
In: \bbtitle{2018 International Conference on Innovations in Science,
  Engineering and Technology (ICISET)},
pp. \bfpage{519}--\blpage{524}
(\byear{2018}).
\doiurl{10.1109/ICISET.2018.8745624}
\end{bchapter}
\endbibitem

\bibitem{bib17}
\begin{bchapter}
\bauthor{\bsnm{Carbonara}, \binits{S.}},
\bauthor{\bsnm{Guaragnella}, \binits{C.}}:
\bctitle{Efficient stairs detection algorithm assisted navigation for vision
  impaired people}.
In: \bbtitle{2014 IEEE International Symposium on Innovations in Intelligent
  Systems and Applications (INISTA) Proceedings},
pp. \bfpage{313}--\blpage{318}
(\byear{2014}).
\doiurl{10.1109/INISTA.2014.6873637}
\end{bchapter}
\endbibitem

\bibitem{bib18}
\begin{bchapter}
\bauthor{\bsnm{Khaliluzzaman}, \binits{M.}},
\bauthor{\bsnm{Deb}, \binits{K.}},
\bauthor{\bsnm{Jo}, \binits{K.-H.}}:
\bctitle{Geometrical feature based stairways detection and recognition using
  depth sensor}.
In: \bbtitle{IECON 2018 - 44th Annual Conference of the IEEE Industrial
  Electronics Society},
pp. \bfpage{3250}--\blpage{3255}
(\byear{2018}).
\doiurl{10.1109/IECON.2018.8591340}
\end{bchapter}
\endbibitem

\bibitem{bib19}
\begin{bchapter}
\bauthor{\bsnm{Huang}, \binits{X.}},
\bauthor{\bsnm{Tang}, \binits{Z.}}:
\bctitle{Staircase detection algorithm based on projection-histogram}.
In: \bbtitle{2018 2nd IEEE Advanced Information
  Management,Communicates,Electronic and Automation Control Conference
  (IMCEC)},
pp. \bfpage{1130}--\blpage{1133}
(\byear{2018}).
\doiurl{10.1109/IMCEC.2018.8469186}
\end{bchapter}
\endbibitem

\bibitem{bib20}
\begin{bchapter}
\bauthor{\bsnm{Redmon}, \binits{J.}},
\bauthor{\bsnm{Divvala}, \binits{S.}},
\bauthor{\bsnm{Girshick}, \binits{R.}},
\bauthor{\bsnm{Farhadi}, \binits{A.}}:
\bctitle{You only look once: Unified, real-time object detection}.
In: \bbtitle{2016 IEEE Conference on Computer Vision and Pattern Recognition
  (CVPR)},
pp. \bfpage{779}--\blpage{788}
(\byear{2016}).
\doiurl{10.1109/CVPR.2016.91}
\end{bchapter}
\endbibitem

\bibitem{bib21}
\begin{bchapter}
\bauthor{\bsnm{Girshick}, \binits{R.}},
\bauthor{\bsnm{Donahue}, \binits{J.}},
\bauthor{\bsnm{Darrell}, \binits{T.}},
\bauthor{\bsnm{Malik}, \binits{J.}}:
\bctitle{Rich feature hierarchies for accurate object detection and semantic
  segmentation}.
In: \bbtitle{2014 IEEE Conference on Computer Vision and Pattern Recognition},
pp. \bfpage{580}--\blpage{587}
(\byear{2014}).
\doiurl{10.1109/CVPR.2014.81}
\end{bchapter}
\endbibitem

\bibitem{bib22}
\begin{bchapter}
\bauthor{\bsnm{Patil}, \binits{U.}},
\bauthor{\bsnm{Gujarathi}, \binits{A.}},
\bauthor{\bsnm{Kulkarni}, \binits{A.}},
\bauthor{\bsnm{Jain}, \binits{A.}},
\bauthor{\bsnm{Malke}, \binits{L.}},
\bauthor{\bsnm{Tekade}, \binits{R.}},
\bauthor{\bsnm{Paigwar}, \binits{K.}},
\bauthor{\bsnm{Chaturvedi}, \binits{P.}}:
\bctitle{Deep learning based stair detection and statistical image filtering
  for autonomous stair climbing}.
In: \bbtitle{2019 Third IEEE International Conference on Robotic Computing
  (IRC)},
pp. \bfpage{159}--\blpage{166}
(\byear{2019}).
\doiurl{10.1109/IRC.2019.00031}
\end{bchapter}
\endbibitem

\bibitem{bib23}
\begin{barticle}
\bauthor{\bsnm{Ilyas}, \binits{M.}},
\bauthor{\bsnm{Lakshmanan}, \binits{A.K.}},
\bauthor{\bsnm{Le}, \binits{A.V.}},
\bauthor{\bsnm{Elara}, \binits{M.R.}}:
\batitle{Staircase recognition and localization using convolution neural
  network (cnn) for cleaning robot application}.
\bjtitle{Preprints 2018}
(\byear{2018}).
\doiurl{10.20944/preprints201812.0296.v1}
\end{barticle}
\endbibitem

\bibitem{bib24}
\begin{bchapter}
\bauthor{\bsnm{Ramteke}, \binits{A.}},
\bauthor{\bsnm{Parabattina}, \binits{B.}},
\bauthor{\bsnm{Das}, \binits{P.K.}}:
\bctitle{A neural network based technique for staircase detection using smart
  phone images}.
In: \bbtitle{2021 Sixth International Conference on Wireless Communications,
  Signal Processing and Networking (WiSPNET)},
pp. \bfpage{374}--\blpage{379}
(\byear{2021}).
\doiurl{10.1109/WiSPNET51692.2021.9419425}
\end{bchapter}
\endbibitem

\bibitem{bib25}
\begin{bchapter}
\bauthor{\bsnm{Ciobanu}, \binits{A.}},
\bauthor{\bsnm{Morar}, \binits{A.}},
\bauthor{\bsnm{Moldoveanu}, \binits{F.}},
\bauthor{\bsnm{Petrescu}, \binits{L.}},
\bauthor{\bsnm{Ferche}, \binits{O.}},
\bauthor{\bsnm{Moldoveanu}, \binits{A.}}:
\bctitle{Real-time indoor staircase detection on mobile devices}.
In: \bbtitle{2017 21st International Conference on Control Systems and Computer
  Science (CSCS)},
pp. \bfpage{287}--\blpage{293}
(\byear{2017}).
\doiurl{10.1109/CSCS.2017.46}
\end{bchapter}
\endbibitem

\bibitem{bib26}
\begin{barticle}
\bauthor{\bsnm{Perez-Yus}, \binits{A.}},
\bauthor{\bsnm{Gutierrez-Gomez}, \binits{D.}},
\bauthor{\bsnm{Lopez-Nicolas}, \binits{G.}},
\bauthor{\bsnm{Guerrero}, \binits{J.J.}}:
\batitle{Stairs detection with odometry-aided traversal from a wearable rgb-d
  camera}.
\bjtitle{Computer Vision and Image Understanding}
\bvolume{154},
\bfpage{192}--\blpage{205}
(\byear{2017}).
\doiurl{10.1016/j.cviu.2016.04.007}
\end{barticle}
\endbibitem

\bibitem{bib27}
\begin{bchapter}
\bauthor{\bsnm{Sinha}, \binits{A.}},
\bauthor{\bsnm{Papadakis}, \binits{P.}},
\bauthor{\bsnm{Elara}, \binits{M.R.}}:
\bctitle{A staircase detection method for 3d point clouds}.
In: \bbtitle{2014 13th International Conference on Control Automation Robotics
  Vision (ICARCV)},
pp. \bfpage{652}--\blpage{656}
(\byear{2014}).
\doiurl{10.1109/ICARCV.2014.7064381}
\end{bchapter}
\endbibitem

\bibitem{bib28}
\begin{botherref}
\oauthor{\bsnm{Tang}, \binits{T.}},
\oauthor{\bsnm{Lui}, \binits{W.}},
\oauthor{\bsnm{Li}, \binits{W.H.}}:
Plane-based detection of staircases using inverse depth.
Australian Robotics and Automation Association,
1--10
(2012)
\end{botherref}
\endbibitem

\bibitem{bib29}
\begin{bchapter}
\bauthor{\bsnm{Stahlschmidt}, \binits{C.}},
\bauthor{\bsnm{Gavriilidis}, \binits{A.}},
\bauthor{\bsnm{Kummert}, \binits{A.}}:
\bctitle{Posture independent stair parameter estimation}.
In: \bbtitle{2015 IEEE International Symposium on Intelligent Control (ISIC)},
pp. \bfpage{65}--\blpage{70}
(\byear{2015}).
\doiurl{10.1109/ISIC.2015.7307281}
\end{bchapter}
\endbibitem

\bibitem{bib30}
\begin{botherref}
\oauthor{\bsnm{Redmon}, \binits{J.}},
\oauthor{\bsnm{Farhadi}, \binits{A.}}:
Yolov3: An incremental improvement
(2018)
{\href{https://arxiv.org/abs/1804.02767}{{arXiv:1804.02767}}}
{[cs.CV]}
\end{botherref}
\endbibitem

\bibitem{bib31}
\begin{botherref}
\oauthor{\bsnm{Ultralytics}}:
YOLOv5,
(2019).
\url{https://github.com/ultralytics/yolov5}
\end{botherref}
\endbibitem

\bibitem{bib32}
\begin{barticle}
\bauthor{\bsnm{Hu}, \binits{J.}},
\bauthor{\bsnm{Shen}, \binits{L.}},
\bauthor{\bsnm{Albanie}, \binits{S.}},
\bauthor{\bsnm{Sun}, \binits{G.}},
\bauthor{\bsnm{Wu}, \binits{E.}}:
\batitle{Squeeze-and-excitation networks}.
\bjtitle{IEEE Transactions on Pattern Analysis and Machine Intelligence}
\bvolume{42}(\bissue{8}),
\bfpage{2011}--\blpage{2023}
(\byear{2020}).
\doiurl{10.1109/TPAMI.2019.2913372}
\end{barticle}
\endbibitem

\bibitem{bib33}
\begin{bchapter}
\bauthor{\bsnm{Xie}, \binits{S.}},
\bauthor{\bsnm{Girshick}, \binits{R.}},
\bauthor{\bsnm{Dollár}, \binits{P.}},
\bauthor{\bsnm{Tu}, \binits{Z.}},
\bauthor{\bsnm{He}, \binits{K.}}:
\bctitle{Aggregated residual transformations for deep neural networks}.
In: \bbtitle{2017 IEEE Conference on Computer Vision and Pattern Recognition
  (CVPR)},
pp. \bfpage{5987}--\blpage{5995}
(\byear{2017}).
\doiurl{10.1109/CVPR.2017.634}
\end{bchapter}
\endbibitem

\bibitem{bib34}
\begin{barticle}
\bauthor{\bsnm{Badrinarayanan}, \binits{V.}},
\bauthor{\bsnm{Kendall}, \binits{A.}},
\bauthor{\bsnm{Cipolla}, \binits{R.}}:
\batitle{Segnet: A deep convolutional encoder-decoder architecture for image
  segmentation}.
\bjtitle{IEEE Transactions on Pattern Analysis and Machine Intelligence}
\bvolume{39}(\bissue{12}),
\bfpage{2481}--\blpage{2495}
(\byear{2017}).
\doiurl{10.1109/TPAMI.2016.2644615}
\end{barticle}
\endbibitem

\bibitem{bib35}
\begin{bchapter}
\bauthor{\bsnm{Redmon}, \binits{J.}},
\bauthor{\bsnm{Farhadi}, \binits{A.}}:
\bctitle{Yolo9000: Better, faster, stronger}.
In: \bbtitle{2017 IEEE Conference on Computer Vision and Pattern Recognition
  (CVPR)},
pp. \bfpage{6517}--\blpage{6525}
(\byear{2017}).
\doiurl{10.1109/CVPR.2017.690}
\end{bchapter}
\endbibitem

\bibitem{bib36}
\begin{bchapter}
\bauthor{\bsnm{He}, \binits{K.}},
\bauthor{\bsnm{Zhang}, \binits{X.}},
\bauthor{\bsnm{Ren}, \binits{S.}},
\bauthor{\bsnm{Sun}, \binits{J.}}:
\bctitle{Deep residual learning for image recognition}.
In: \bbtitle{2016 IEEE Conference on Computer Vision and Pattern Recognition
  (CVPR)},
pp. \bfpage{770}--\blpage{778}
(\byear{2016}).
\doiurl{10.1109/CVPR.2016.90}
\end{bchapter}
\endbibitem

\bibitem{bib37}
\begin{botherref}
\oauthor{\bsnm{Paszke}, \binits{A.}},
\oauthor{\bsnm{Chaurasia}, \binits{A.}},
\oauthor{\bsnm{Kim}, \binits{S.}},
\oauthor{\bsnm{Culurciello}, \binits{E.}}:
Enet: A deep neural network architecture for real-time semantic segmentation
(2016)
{\href{https://arxiv.org/abs/1606.02147}{{arXiv:1606.02147}}}
{[cs.CV]}
\end{botherref}
\endbibitem

\bibitem{bib38}
\begin{botherref}
\oauthor{\bsnm{Garcia-Garcia}, \binits{A.}},
\oauthor{\bsnm{Orts-Escolano}, \binits{S.}},
\oauthor{\bsnm{Oprea}, \binits{S.}},
\oauthor{\bsnm{Villena-Martinez}, \binits{V.}},
\oauthor{\bsnm{Garcia-Rodriguez}, \binits{J.}}:
A review on deep learning techniques applied to semantic segmentation
(2017)
{\href{https://arxiv.org/abs/1704.06857}{{arXiv:1704.06857}}}
{[cs.CV]}
\end{botherref}
\endbibitem

\bibitem{bib39}
\begin{bchapter}
\bauthor{\bsnm{Tekin}, \binits{B.}},
\bauthor{\bsnm{Sinha}, \binits{S.N.}},
\bauthor{\bsnm{Fua}, \binits{P.}}:
\bctitle{Real-time seamless single shot 6d object pose prediction}.
In: \bbtitle{2018 IEEE/CVF Conference on Computer Vision and Pattern
  Recognition},
pp. \bfpage{292}--\blpage{301}
(\byear{2018}).
\doiurl{10.1109/CVPR.2018.00038}
\end{bchapter}
\endbibitem

\end{thebibliography}


\end{document}